\documentclass[preprint,5p,twocolumn]{elsarticle}

\journal{Neural Networks}
\usepackage{natbib}
\renewcommand{\cite}{\citep}
\usepackage{amsfonts}
\usepackage{amsmath}
\usepackage{amssymb}
\usepackage{amstext}
\usepackage{bm}

\usepackage{epsf,graphicx}
\graphicspath{{./Figures/}}

\usepackage{url}          %%% Permite cortar bien las URLs al final de linea

\def\vec#1{\mbox{\boldmath$\displaystyle#1$}}
\def\E    {\displaystyle{e}}
\def\b    {\vec{b}}
\def\c    {\vec{c}}
\def\W    {\vec{W}}
\def\Wi   {\W_{\!i\ }}
\def\h    {\vec{h}}
\def\x    {\vec{x}}
\def\xx   {\stackrel{\thicksim}{\vec{x}}}
\def\y    {\vec{y}}

\def\lgst {\text{lgst}} %%% Logistic function

\begin{document}

\title{Weighted Contrastive Divergence}

\author[cs]{Enrique~Romero~Merino}
\ead{eromero@cs.upc.edu}
\author[fen]{Ferran~Mazzanti~Castrillejo}
\ead{ferran.mazzanti@upc.edu}
\author[cs]{Jordi~Delgado~Pin}
\ead{jdelgado@cs.upc.edu}
\author[bsc]{David~Buchaca~Prats}
\ead{david.buchaca@bsc.es}
\address[cs]{Departament de Ci\`encies de la Computaci\'o, Universitat
  Polit\`ecnica de Catalunya - BarcelonaTech, Spain}
\address[fen]{Departament de F\'{\i}sica i Enginyeria Nuclear,
  Universitat Polit\`ecnica de Catalunya - BarcelonaTech, Spain}
\address[bsc]{Barcelona Supercomputing Center (BSC), 
  Universitat Polit\`ecnica de Catalunya - BarcelonaTech, Spain}

\begin{abstract}
  Learning algorithms for energy based Boltzmann architectures that
  rely on gradient descent are in general computationally prohibitive,
  typically due to the exponential number of terms involved in
  computing the partition function. In this way one has to resort to
  approximation schemes for the evaluation of the gradient.  This is
  the case of Restricted Boltzmann Machines (RBM) and its learning
  algorithm Contrastive Divergence (CD). It is well-known that CD has
  a number of shortcomings, and its approximation to the gradient has
  several drawbacks. Overcoming these defects has been the basis of
  much research and new algorithms have been devised, such as
  persistent CD. In this manuscript we propose a new algorithm that we
  call {\em Weighted CD} (WCD), built from small modifications of the
  negative phase in standard CD. However small these modifications may
  be, experimental work reported in this paper suggest that WCD
  provides a significant improvement over standard CD and persistent
  CD at a small additional computational cost.
\end{abstract}

%\begin{keywords}
%Neural networks, Restricted Boltzmann Machine, Contrastive Divergence
%\end{keywords}

\maketitle

\section{Introduction}

Restricted Boltzmann Machines (RBM), originally conceived in the
eighties (called {\em
  harmonium}~\cite{smolensky-Restricted-Boltzmann-Machines-1986-PDP})
as a topological simplification of Boltzmann Machines~(BM), have
captured the attention of the neural network community in the last
decade. This is because of its role as building blocks of multilayer
learning architectures such as Deep Belief Networks
(DBN)~\cite{hinton-et-al-DeepBeliefNetworks-2006-NC} or deep
autoencoders~\cite{hinton-salakhutdinov-Auto-Encoders-2006-Science}. RBMs
have been successfully applied in several areas of interest, such as
image
classification~\cite{hinton-salakhutdinov-Auto-Encoders-2006-Science},
collaborative filtering~\cite{salakhutdinov2007restricted} or acoustic
modeling~\cite{mohamed2012acoustic} to mention only a few.

An RBM is able to learn a target probability distribution from
samples.  RBMs have two layers, one of {\em hidden} and another of
{\em visible} units, and no intra-layer connections. This property
makes working with RBMs simpler than with regular BMs. This is because
the stochastic computation of the log-likelihood gradient may be more
efficiently evaluated, since Gibbs
sampling~\cite{bengio-DeepArchitectures-2009-bk} can be performed in
parallel.  Similar to BMs, RBM are universal
approximators~\cite{leroux-bengio-Representational-Power-RBMs-2008-NC},
in the sense that, given enough hidden units, RBMs can approximate any
probability distribution.

In 2002, the \textit{Contrastive Divergence} learning algorithm (CD)
was put forward as an efficient training method for product-of-expert
models, from which RBMs are a special
case~\cite{hinton-Contrastive-Divergence-2002-NC}. It was observed
that using CD to train RBMs worked quite well in practice. This fact
is important for deep learning with RBMs since some authors have
suggested that a multi-layer deep neural network is better trained
when each layer is separately pre-trained, as if it were a single
RBM~\cite{hinton-salakhutdinov-Auto-Encoders-2006-Science,
  bengio-et-al-ContinuousInputs-And-AutoEncoders-2007-NIPS,
  larochelle-et-al-Strategies-Train-Deep-Networks-2009-JMLR}. Thus,
training RBMs with CD and stacking up RBMs is a possible way to go
when designing deep learning architectures. In any case, the
probabilistic potential of the RBM has been largely overlooked.  More
recently, RBMs have found interesting applications in solving
challenging problems that are otherwise very difficult to
tackle~\cite{Troyer_2017}.

Contrastive Divergence is an approximation to the true, but
computationally intractable, RBM log-likelihood
gradient~\cite{bengio-delalleau-Justifying-ContrastiveDivergence-2009-NC,
  fischer-igel-Bounding-Bias-CD-2011-NC}. As such, it is far from
being perfect: It is biased and it may not even
converge~\cite{carreira-hinton-Contrastive-Divergence-Learning-2005-AISTATS,
  yuille-Convergence-Contrastive-Divergence-2005-NIPS,
  mackay-Failures-Contrastive-Divergence-2001-TR}. Also CD, and
variants such as Persistent CD
(PCD)~\cite{tieleman-Persistent-Contrastive-Divergence-2008-ICML} or
Fast Persistent
CD~\cite{tieleman-hinton-Improved-Persistent-Contrastive-Divergence-2009-ICML}
can lead to a steady decrease of the log-likelihood during
learning~\cite{fischer-igel-Divergence-Contrastive-Divergence-2010-ICANN,
  desjardins-et-al-Parallel-Tempering-2010-AISTATS}. Furthermore, the
% models obtained at the point of maximum log-likelihood (which should
% be detected with a reliable stopping criterion, see~\cite{PEAZO-AUTOCITA})
maximum log-likelihood models are such that the learned probability
distribution accumulates most of the probability mass only on a small
number of states.
%% ~\cite{???A QUIEN? quizá
%%   schulz-et-al-Convergence-Contrastive-Divergence-2010-NIPSw}.
%% {\bf ??? incluir referencia}

% the probability mass of the states learnt
% by the RBM
% % , those with significative probability,
% is not distributed
% among all those states according to the target distribution, but it
% accumulates on a few states

In this paper we propose an alternative approximation to the CD
gradient called {\em Weighted Contrastive Divergence} (WCD). The main
difference consists in weighting the elements involved in the negative
phase by its relative probability in the batch.  This small but
significant change leads to probability distributions that more
closely resemble the target ones, at the expense of a 
% with a non-degenerating
% log-likelihood. Moreover,
% and remarkably,
not very large additional computational cost.
% is almost negligible when compared to standard CD.
In order to illustrate these points explicitly, we
address small size problems that allow for an exact evaluation of both
the partition function and the Kullback-Leibler divergence, which is
directly related to the log-likelihood of the data.  In this way we
can compare the target and model probabilities of each state, thus
showing the benefits of the scheme proposed at a detailed level.
We also analyze real-world large size problems, 
% We omit on purpose large problems
where the partition function can not
be evaluated.
% as we want to compare exact probabilities.
Due to the intractability of an exact calculation of the probabilities, 
we use a Parzen window estimator~\cite{Breuleux_Parzen_2011} to
measure the quality of the results, finding that WCD provides
a significant improvement in the resulting model.
A similar approach but with a different weighting scheme has recently
appeared in~\cite{Krause_2018}.

The paper is organized as follows.  Section~\ref{section-Standard-CD}
reviews the RBM model and CD. In section~\ref{section_WCD} we present
the basic WCD algorithm and its natural extension, Persistent
Weighted CD (WPCD).  Experiments
showing the benefits of WCD are described and presented in
sections~\ref{section-experiments} and~\ref{section-experiments-Parzen}.

% propose a few modifications of standard CD after reviewing in detail
% its shortcomings, to devise WCD, a new learning algorithm for
% RBMs. The modifications introduced in standard CD are not very
% complicated nor sophisticated, but the improvement over standard CD is
% quite remarkable, as suggested by the experiments carried out (results
% detailed in section~\ref{section-experiments}).

\section{Standard Contrastive Divergence\label{section-Standard-CD}}

Binary Restricted Boltzmann Machines are energy-based
probabilistic models whose energy function is:
\begin{equation}
\label{energy-RBM-discrete-binary-binary}
 \text{Energy}(\x,\h) = -\b^t\x - \c^t\h - \h^t\W\x \ ,
\end{equation}
where $\x$ and $\h$ are binary visible and binary hidden variables,
respectively. Hidden variables are usually introduced to increase the
expressive power of the model. The probability distribution of the
visible variables is defined as the marginal distribution
\begin{equation}
\label{pdf-energy-x-sumh}
 P(\x) = \frac{\sum_{\h} \E^{-\text{Energy}(\x,\h)}}{Z} \ ,
\end{equation}
in terms of the partition function 
\begin{equation}
 Z = \sum_{\x,\h} \E^{-\text{Energy}(\x,\h)} \ .
\end{equation}
The particular form of the energy function allows to efficiently
compute the Free Energy in the numerator $e^{-\text{FreeEnergy}(\x)} = \sum_{\h}
\E^{-\text{Energy}(\x,\h)}$ of Eq.~(\ref{pdf-energy-x-sumh}). In addition,
since both $P(\h|\x)$ and $P(\x|\h)$ factorize, it is possible to
compute $P(\h|\x)$ and $P(\x|\h)$ in one step, making possible to
perform Gibbs sampling efficiently
\cite{geman-geman-Gibbs-Sampling-1984-TPAMI}.  However, the evaluation
of $Z$ is still computationally prohibitive when the number of input
and hidden variables is large
%since it involves an exponentially large number of terms
%In fact, the partition function cannot be computed efficiently
\cite{long-serveido-RBMs-Hard-Evaluate-Simulate-2010-ICML}.

The energy function depends on several parameters $\theta =
\{\b,\c,\W\}$. Given a data set $X=\{\x^1,\dots,\x^N\}$, learning with
RBMs consists in adjusting $\theta$ so as to
maximize the log-likelihood of the data. In energy-based models, the
derivative of the log-likelihood can be expressed as
%\begin{align}
\begin{eqnarray}
\label{dlog-likelihood-Exact}
\lefteqn{\!\!\!\!\!
 -\frac{\partial\log P(\x;\theta)}{\partial\theta} =
  \ E_{P(\h|\x)} \left[\frac{
      \partial\text{Energy}(\x,\h)}{\partial\theta}\right]} \nonumber
\\
{} & \ \ \ \ \ \ \ \ \ \ \ \ \ \ \ \
-\ E_{P(\xx)} \left[E_{P(\h|\xx)}\left[\frac{\partial\text{Energy}(\xx,\h)}{\partial\theta}\right] \right]
\end{eqnarray}
%\end{align}
where the first term is called the {\em positive phase} and the second
term the {\em negative phase}.  Similar to (\ref{pdf-energy-x-sumh}),
the exact computation of the derivative of the log-likelihood is in
general computationally prohibitive because the negative phase
in~(\ref{dlog-likelihood-Exact}) can not be efficiently computed.
This is due to the fact that the negative phase comes from the
derivative of the logarithm of the partition function.  Notice that
the log-likelihood of the data and the Kullback-Leibler (KL)
divergence contain the same information and can be used
indistinguishably~\cite{coolen2005theory}.

The most common learning algorithm for RBMs is called {\em Contrastive
  Divergence} (CD)~\cite{hinton-Contrastive-Divergence-2002-NC}.  The
algorithm for CD$_k$ estimates the derivative of the log-likelihood as
%\begin{align}
\begin{eqnarray}
\label{dlog-likelihood-CDk}
\lefteqn{-\frac{\partial\log P(\x;\theta)}{\partial\theta} \simeq
  \ E_{P(\h|\x)} \left[\frac{
      \partial\text{Energy}(\x,\h)}{\partial\theta}\right]} \nonumber
\\
{} & \ \ \ \ \ \ \ \ \ \ \ \ \ \ \ \ \ \ \ \
-\ E_{P(\h|\x_{k})}\left[\frac{\partial\text{Energy}(\x_{k},\h)}{\partial\theta}\right]
\end{eqnarray}
%\end{align}
where $\x_{k}$ is the last sample from the Gibbs chain starting
from $\x$ obtained after $k$ steps
\begin{itemize}
\item[] $\h_1 \sim P(\h|\x)$
\item[] $\x_1 \sim P(\x|\h_1)$
\item[] ...
\item[] $\h_k \sim P(\h|\x_{k-1})$
\item[] $\x_{k} \sim P(\x|\h_k)$ \ .
\end{itemize}

For binary RBMs,
$E_{P(\h|\x)}\left[\frac{\partial\text{Energy}(\x,\h)}{\partial\theta}\right]$
can be easily computed and yields
\begin{itemize}
\item $E_{P(\h|\x)}
  \left[\frac{\partial\text{Energy}(\x,\h)}{\partial\b_j}\right] = -\x_j$
\item $E_{P(\h|\x)}
  \left[\frac{\partial\text{Energy}(\x,\h)}{\partial\c_i}\right] = -\lgst\left(\c_i + \Wi\x\right)$
\item $E_{P(\h|\x)}
  \left[\frac{\partial\text{Energy}(\x,\h)}{\partial\W_{ij}}\right] = -\x_j \cdot \lgst\left(\c_i + \Wi\x\right)$
\end{itemize}
where $\Wi$ is the $i$-th row of $\W$ and $\lgst(z)$ stands for the logistic
function $\lgst(z) = \frac{1}{1+\E^{-z}}$.

With these definitions, the modification of the weights with standard
CD$_k$
% gradient ascent
for a training set $X=\{\x^1,\dots,\x^N\}$ reads
% and estimating the derivative of the log-likelhood with standard
% CD$_k$ is
\begin{eqnarray}
  \Delta(\theta)  \!\!\! & = & \!\!\!
     \frac{1}{N} \sum_{i=1}^{N}
     E_{P(\h|\x)} \left[\frac{
       \partial\text{Energy}(\x,\h)}{\partial\theta}\right]
  \nonumber \\ [2mm]
  & & \!\!\! 
-\frac{1}{N} \sum_{i=1}^{N}
     E_{P(\h|\x_{k})}\left[\frac{\partial\text{Energy}(\x_{k},\h)} 
       {\partial\theta}\right] \ .
\label{delta_theta}     
\end{eqnarray}
  
Notice that the factor $1/N$ weights equally every example in the
training set, while the different probability each state should get
comes from the repetition of the examples. This is important when the
probabilities to be learned are non-uniform. 

Nowadays the standard CD algorithm is hardly used in favor of its
improved version PCD, where a continuous Markov chain is used to
sample the negative
phase~\cite{tieleman-Persistent-Contrastive-Divergence-2008-ICML}.

\section{Weighted Contrastive Divergence}
\label{section_WCD}

In this section we describe the modification to the family of CD
algorithms proposed in this work, that we generically call {\em
  Weighted Contrastive Divergence} (WCD). First we point out the main
limitations of CD, then we provide the description of the WCD
algorithm, and finally present its extension to its Persistent
version. 

\subsection{Non-desirable Behavior of Standard CD$_k$ \label{section-behavior-CDk}}

As previously mentioned, the exact evaluation of the partition
function is in general not possible.  Therefore, it is difficult to
compare the behavior of CD$_k$ in real world problems with respect to
the exact gradient in Eq.~(\ref{dlog-likelihood-Exact}).  However, it
is expected that the conclusions drawn for small problems, where all
the probabilities can be exactly computed, may be extrapolated to
large ones.

We performed an extensive evaluation of standard CD$_k$ in problems
with a tractable number of input units (see
section~\ref{section-experiments}). From these experiments we can
conclude that, in practice, standard CD$_k$ shows the following
properties:
\begin{enumerate}

\item In many cases, standard CD$_k$ is not able to obtain good models
  for any combination of parameters, i.e, it is not able to assign
  large enough probabilities to the examples in the training set. This
  effect is more noticeable when $k$ is small or the number of hidden
  units is small. In some cases, it happens even when $k$ or the
  number of hidden units is large (see table~\ref{table_KL}),
  %{table_datasets}), 
  %(see figure \ref{fig_01-StandardCD-NoLearning}),
    and it seems
  related to the difficulty of the problem.

  %\\ *** LS11 (CD01 con pocas hidden units)
  %\\ *** LS15 (CD01)
  %\\ *** P08  (CD01)
  %\\ *** P10  (CD01 / CD10 con pocas hidden units)

\item 
  For good combination of parameters, standard CD$_k$ is able to
  obtain somewhat good models, preserving the sum of probabilities of
  the training set. However, in many cases    
%  the probabilities in the training set
%  sum up very quickly to values very close to $1$. However, in many
%  problems where all states in the training set should receive the
%  same probability,
  it tends to concentrate most of the probability mass in a few
  states, even in cases where all states in the training set should
  receive the same probability.
  %% For instance, the blue curves in
  %% figure~\ref{fig_03-Compare-StandardCD-WeightedCD-02} show that the
  %% difference between the maximum and minimum probabilities in the
  %% training set increases as learning evolves.
  As a consequence, the
  likelihood presents a non-monotonic behavior: it starts increasing,
  then reaches a maximum and starts to decrease. Accordingly, the
  Kullback-Leibler (KL) divergence starts decreasing, reaches a
  minimum and then increases. This behaviour can be seen in several
  figures of section~\ref{section-experiments}).

\end{enumerate}

The idea behind the WCD algorithm is to (at least partially) overcome
these limitations. 

\subsection{Description of WCD \label{section-weighted-contrastive-divergence}}

Since the positive phase of standard CD$_k$ is exactly equal to the
positive phase of the exact gradient, there is no need to modify
it. In contrast, the negative phase in CD$_k$ suffers from extreme and
drastic approximations that are responsible for the limitations
described above. For that reason, we propose a modification of the
negative phase of CD$_k$. This modification consists in weighting
differently every contributing state in the negative phase.  We call
this new algorithm {\em Weighted Contrastive Divergence}, which we
describe in the following.

The negative phase of the exact gradient from
Eq.~(\ref{dlog-likelihood-Exact}) reads
% with stochastic gradient ascent is
\begin{equation}
\label{negative-phase-Exact-minibatch-expanded}
  \sum_{\xx} \ P(\xx)
  \ E_{P(\h|\xx)}\left[\frac{\partial\text{Energy}(\xx,\h)}{\partial\theta}\right]
  \ ,
\end{equation}
and depends on all states in the space.
If we consider, as usual in practice, optimization with stochastic
gradient ascent, the negative phase proposed by CD$_k$ is
\begin{equation}
\label{negative-phase-CDk-minibatch}
  \sum_{i=1}^{N_B} \ \frac{1}{N_B}
  \ E_{P(\h|\x_{k}^i)}\left[\frac{\partial\text{Energy}(\x_{k}^i,\h)}{\partial\theta}\right]
\end{equation}
where $N_B$ is the number of examples in the batch.  In this
expression, $\x_k^i$ stands for the $k$-th step Gibbs-sampling
reconstruction of the $i$-th element $\x^i$ of the batch.
%% Equation (\ref{negative-phase-CDk-minibatch}) is related to the Monte
%% Carlo Markov chain that should be run in order to obtain an accurate
%% approximation of the negative phase.
For small values of $N_B$ and $k$ (which is usually the case), it
usually is a very rough approximation.

There are several important differences between the respective
negatives phases in~(\ref{negative-phase-Exact-minibatch-expanded})
and~(\ref{negative-phase-CDk-minibatch}):
\begin{enumerate}
\item CD$_k$ computes the sum over the reconstructions of the data,
  whereas the exact gradient computes the sum over the whole space
  (which includes the reconstructions of the data). It means that
  CD$_k$ only explores, for every batch, a tiny fraction of the
  space. This is good from the efficiency point of view, but bad from
  the statistical side. 
\item CD$_k$ weights every element in the sum by a weighting factor
  that comes from the k-step sampling distributions. However, for
  small $k$, large space size and modest batch size, repetitions in
  the reconstructions are highly unlikely and the implemented
  weighting factor is nearly the constant $\frac{1}{N_B}$, whereas the
  exact gradient weights every element by the model probability
  $P(\xx)$. It means that CD$_k$ for low $k$ approximately gives the
  same weight to every element, in contrast to the exact gradient that
  gives more importance to elements with larger probabilities.
\end{enumerate}

The WCD algorithm proposes a way to overcome the second issue.
% Overcoming the first issue is in general not possible because 
% the computation of (\ref{negative-phase-Exact-minibatch-expanded}) is
% prohibitive as it involves an exponentially large number of terms.
% % , involving a sum of $2^{N_v}$ terms where
% % $N_v$ is the input dimension.
% Overcoming the second difference also presents important difficulties,
% since the computation of $P(\xx)$ involves the computation of the
% partition function, which is also computationally
% prohibitive. However,
One can assign larger weights to elements with larger probabilities by
computing the relative probabilities of the elements in the batch
\begin{equation}
\label{weights-negative-phase-probs}
 \overline{P}(\x_k^i) = \frac{P(\x_k^i)}{\sum_{j=1}^{N_B} P(\x_k^j)} \ ,
\end{equation}
where, as usual, $\x_{k}^i$ is the last sample from the Gibbs chain
starting from $\x^i$ and obtained after $k$ steps, and $N_B$ is the number
of examples in the batch.
In this way the proposed negative phase in WCD becomes
% , which
% can be obtained by computing the numerator in
% Eq.~(\ref{pdf-energy-x-sumh}) and normalizing them.
\begin{equation}
\label{negative-phase-WeightedCDk-minibatch}
  \sum_{i=1}^{N_B} \ \overline{P}(\x_{k}^i)
  \ E_{P(\h|\x_{k}^i)}\left[\frac{\partial\text{Energy}(\x_{k}^i,\h)}{\partial\theta}\right]
  \ .
\end{equation}
% (and maintain exactly the relative contributions of every element)

Every weight $\overline{P}(\x^i_k)$ in
(\ref{weights-negative-phase-probs}) can be efficiently computed.
First, because the partition function $Z$ cancels out. Second, because
Eq.~(\ref{weights-negative-phase-probs}) is equivalent to
\begin{equation}
\label{weights-negative-phase-free-energy}
\overline{P}(\x_k^i) = \frac{\E^{\text{-FreeEnergy}(\x_k^i)}}
         {\sum_{j=1}^{N_B} \E^{\text{-FreeEnergy}(\x_k^j)}}
\end{equation}
while the Free Energy factorizes in the RBM topology, as previously
mentioned.  To summarize, in WCD one evaluates the negative phase as
described in Eq.~(\ref{negative-phase-WeightedCDk-minibatch}), while
the positive phase is the same as in standard CD.

Intuitively, weighting the negative phase as in
Eq.~(\ref{negative-phase-WeightedCDk-minibatch})
allows to obtain better estimators of the real negative phase
in Eq.~(\ref{negative-phase-Exact-minibatch-expanded}),
which also weights differently every state, assigning more weight
to the states that have a larger Boltzmann
probability. 
%% This is because the actual negative phase is an statistical average
%% over the whole space with a Boltzmann probability that differs from one
%% state to another.
As we will confirm in the experiments with small problems, this
modification has a positive effect on the learning process, allowing
to obtain models with much lower KL values than standard CD$_k$. In
addition, and differently from standard CD$_k$, the proposed approach
fits better the training probability distribution as it is shown in
section~\ref{section-experiments}.  Moreover, the KL has a
decreasing behavior during learning.  Therefore, weighting the
negative phase helps overcoming the non-desired behavior of standard
CD$_k$ pointed out in section~\ref{section-behavior-CDk}.
Even though such a detailed study is unfeasible with real-world large
problems, the approach employed in
section~\ref{section-experiments-Parzen} seems to indicate that the
use of a weighted negative phase also improves the statistical
representativity of the model.

\subsection{Generalization to Weighted Negative Phase}

One can easily generalize the previous procedure to any variant of
standard CD. To do so, one changes the reconstructions of the data in
Eq.~(\ref{negative-phase-WeightedCDk-minibatch}) by a 
suitable choice of a set of points $Y=\{\y^1,
\y^2, \ldots, \y^M\}$, leading to the {\em Weighted Negative Phase},
defined as
\begin{equation}
\label{weighted-negative-phase_generalized}
  \sum_{i=1}^{M} \ \overline{P}(\y^i)
  \ E_{P(\h|\y^i)}\left[\frac{\partial\text{Energy}(\y^i,\h)}{\partial\theta}\right]
  \ .
\end{equation}
The previous definition does not impose any condition on the subset
$Y$. This gives a lot of flexibility, but it also presents the
additional problem of selecting a good set of candidates.  In
principle, any subset could be used. For instance, in the special but
relevant case of PCD, $Y$ can be taken as the set of persistent
reconstructions of the data. We will denote its corresponding Weighted
version as {\em Weighted Persistent CD} (WPCD).  On the other hand, if
$Y$ spans the whole space,
Eq.~(\ref{weighted-negative-phase_generalized}) is the negative phase
of the exact gradient. Obviously, the computational cost is directly
proportional to the number of elements in $Y$.  Notice that, in
general the computational overhead associated to the calculation of
the negative phase in Eq.~(\ref{weighted-negative-phase_generalized})
is small compared with the one corresponding to the evaluation in
standard CD$_k$, although this obviously depends on $Y$.

\section{Experiments with small size problems  \label{section-experiments}}

In the following we perform a series of experiments to test the
proposed approach. More precisely, we compare standard CD$_k$ for
$k=1$ and $k=10$, together with PCD, to their {\em Weighted}
counterparts, WCD$_k$ and WPCD. In this section we restrict the
analysis to small dimensional spaces where exact calculations can be
performed. Our goal is to compare the different models at the lowest
possible level and to avoid drawing conclusions from a coarse
approximation, as usually done when dealing with larger problems.  In
particular we evaluate the exact partition function of each model,
compute the exact probabilities of the whole space, and compare the
probabilities of the data in the different models.  We also evaluate
the exact KL of the obtained models.

\subsection{Data Sets}

We have tested the proposed approach in a series of data sets
described in the following. Training is performed including examples
and probabilities. We will denote {\em target distribution} the set of
probabilities assigned to all the states in each data set.  Three
different schemes have been used to establish the target
distributions.  The simplest one is the empirical distribution, that
sets the same (uniform) probability to each state. The second one
% ,$G(\mu,\sigma)$,
draws the probabilities from a Gaussian profile.
% of mean $\mu$ and standard deviation $\sigma$.
The third model assigns
different but uniform probabilities to separate subsets.  We will call
{\em training space} to the combination of data and target
distribution associated to the data.

The first problem, denoted {\em Bars and Stripes}, consists in
detecting vertical and horizontal lines in binary images containing
either of them but not both.  Two versions of this problem were
tested, containing $3\times 3$ (BS09) or $4\times 4$ (BS16) images,
respectively. The second problem, named {\em Labeled Shifter Ensemble}
(LSE), consists in learning a number of states formed as follows:
given an initial $N$-bit pattern, generate three new states
concatenating to it the bit sequences 001, 010 or 100 and a new
$N$-bit pattern computed as the original one shifting one bit to the
left if the intermediate code is 001, copying it unchanged if the code
is 010, or shifting it one bit to the right if the code is 100. The
size of the states are $2N+3$ bits. Again, two versions of this
problem were evaluated, with $N=4$ (LSE11) and $N=6$ (LSE15).  These
problems have already been explored in
\cite{fischer-igel-Divergence-Contrastive-Divergence-2010-ICANN}.

The third data set tested is the Parity problem, which consists in
learning whether the number of bits with value 1 is even or not.  The
Parity problem is known to be very difficult to learn with classical
neural
networks~\cite{rumelhart-et-al-1986,bengio-lecun-Scaling-Learning-AI-2007-bookCh}.
It is easy to understand that this is a very difficult problem to
learn in the context of Boltzmann Machines as in a high order model it
requires a single weight connecting all units
simultaneously~\cite{farguell2008boltzmann}. This problem was tested
with $8$ and $10$ input variables (P08 and P10, respectively).

The target distributions associated to the BS09, BS16, LSE11, LSE15,
P08 and P10 problems are the corresponding empirical
distributions. That is, every element in each data set has a probability
equal to one over the number of elements in the data set.

The fourth tested problem, which we call Int12, is a data set
containing $N=2^{12}$ integers.  The unnormalized probability assigned
to the each integer $n\in\{0,1,2\ldots, 2^{12}-1\}$ 
is given by the following expression
\begin{equation}
\label{eq:gaussian_like}
q(n) = p_{max} e^{- \lambda n^2} \ ,
\end{equation}
% where ${\mathcal N}$ is a normalization constant and
where $\lambda$ is 
a parameter that depends on the maximum and minimum probabilities in
the data set, $p_{max}$ and $p_{min}$. 
The probability assigned to each element in the data set is 
computed as follows:
\begin{equation}
p(n)={ q(n) \over \sum_{m=0}^N q(m) } \ ,
\label{Gauss}
\end{equation}
where  
% By choosing a maximum and mininum probability for any state $Pmax$ and
% $ Pmin$ values and assigning $\lambda$ to be
\[
\lambda = \frac{1}{(N-1)^2} \ln\left( \frac{p_{min}}{p_{max}} \right) \ .
\]

The last two data sets assign different probabilities to the same
$2^{12}$ integers, formed by the ordered list
$[0,3,6,\ldots,1,4,7,\ldots,2,5,8,\ldots]$. 
%% The last two datasets have the same elements as the Int12
%% one, which are two variants of this problem and both of them make
%% the same permutation to the dataset.  This permutation orders all
%% integers by first setting the multiples of 3, $\dot{3}$, then $\dot{3}
%% =1$ and then $\dot{3}+2$.
The first variant, Mult3G, assigns to each
position in the list the probability defined by
Eq.~(\ref{Gauss}). The second variant, Mult3D, assigns three different 
probability values to the elements in the list that belong to the
$\dot 3$, $\dot 3 + 1$ and $\dot 3 + 2$ sublists, respectively.
These values are fixed imposing the sum of the probabilities in
each group to be $0.6, 0.3$ and $0.1$. We denote this scheme as {\em
  Discrete}. 
Table~\ref{table_datasets} summarizes the main properties of the
training spaces used in the experiments.

\begin{table*}[t!]
\begin{center}
{
\begin{tabular}{| lccl |} \hline
Data set & Input dimension & Data set size & Target distribution 
\\ \hline
BS09      &   9  &    14  &  Empirical \\
BS16      &  16  &    30  &  Empirical \\
LSE11     &  11  &    48  &  Empirical \\
LSE15     &  15  &   192  &  Empirical \\
P08       &   8  &   128  &  Empirical \\
P10       &  10  &   512  &  Empirical \\
Int12     &  12  &  4096  &  Gaussian  Profile \\
Mult3G    &  12  &  4096  &  Gaussian  Profile \\
Mult3D    &  12  &  4096  &  Discrete  \\
 \hline 
\end{tabular}
\caption{Description of the different data sets used}
\label{table_datasets}
}
\end{center}
\end{table*}

\subsection{Experimental Setting}

%%% Plantearse si incluir CD0
%
%%% CDkWeighted_RatioRand02p0_KeepGood0_k00
%%% CDkWeighted_RatioRand01p0_KeepGood1_k00
%
% Consider that, for $k=0$, CD$_0$ ***definir-bien*** the
% reconstructions obtained by CD$_0$ are computed with the identity
% function (i.e., the reconstruction of an example is the example
% itself).
%
% In addition to random examples, $Y$ may also contain the examples in
% the training set. --- justificar
% We will call this model {\it Weighted Negative Phase with Random
%   Examples and Training Data} (WNPRETD). Note that the training data
% in the negative phase do not cancel out the positive phase, since in
% the negative phase the elements are weighted by their relative
% probability.
%
% WNPRETD is a particular case of WCD$_k$RE where the reconstructions of
% the data are the data themselves and no random example is added.
%
% \item WCD$_0$RE (equivalent to WNPRETD): $Y$ contains the training
%  data and random examples
%  \\ (WeightedCD00-RatioRandN-KeepGood1)
%
% Para CD0 hay que ajustar muy bien los parametros de aprendizaje

The experiments were performed in two steps. In the first one we
selected, for every data set, suitable parameters for standard CD$_k$
($k=1$ and $k=10$) and standard PCD. This selection was performed
independently for every model. In the second step, we used the
parameters found in the first step to test WCD$_k$ and WPCD.

\begin{table*}[t!]
\begin{center}
{
\def\arraystretch{1.3}
\begin{tabular}{|ccccccc|} \hline
CD$_1$ & CD$_{10}$ & PCD & WCD$_1$ & WCD$_{10}$ & WPCD & Data set
\\ \hline
0.0450 (0.0206) & 0.0035 (0.0013) & 0.0962 (0.0052) & 0.0011 (0.0001) & 0.0011 (0.0001) & 0.0464 (0.0088) & BS09  \\
0.1657 (0.0578) & 0.0212 (0.0077) & 0.1850 (0.0062) & 0.0008 (0.0001) & 0.0007 (0.0001) & 0.1060 (0.0141) & BS16  \\
0.2986 (0.0641) & 0.0634 (0.0274) & 0.1918 (0.0172) & 0.1043 (0.0363) & 0.0113 (0.0024) & 0.1050 (0.0070) & LSE11  \\
0.7767 (0.0913) & 0.2194 (0.0441) & 0.2379 (0.0136) & 0.3558 (0.0888) & 0.0233 (0.0021) & 0.1822 (0.0071) & LSE15  \\
0.6464 (0.0858) & 0.1530 (0.0435) & 0.2681 (0.0591) & 0.6535 (0.0791) & 0.0201 (0.0043) & 0.1515 (0.0189) & P08  \\
0.6933 (0.0001) & 0.3062 (0.0169) & 0.1810 (0.0456) & 0.6937 (0.0019) & 0.0681 (0.0232) & 0.1638 (0.0313) & P10  \\
0.0221 (0.0262) & 0.0007 (0.0014) & 0.0220 (0.0003) & 0.0022 (0.0034) & 0.0021 (0.0025) & 0.0220 (0.0003) & Int12   \\ 
0.5453 (0.0001) & 0.0220 (0.0009) & 0.1778 (0.0336) & 0.0041 (0.0009) & 0.0064 (0.0012) & 0.1960 (0.0948) & Mult3G  \\
0.4246 (0.0001) & 0.3120 (0.1113) & 0.1958 (0.0250) & 0.0036 (0.0007) & 0.0052 (0.0013) & 0.2025 (0.0206) & Mult3D  \\ \hline 
\end{tabular}
\caption{Mean KL values for the different problems
  described in the text. The standard deviations are in parenthesis.}
\label{table_KL}
}
\end{center}
\end{table*}

%\begin{table*}[t!]
%\begin{center}
%{
%% \tabcolsep=4mm
%\def\arraystretch{1.3}
%\begin{tabular}{|ccccccc|} \hline
%CD$_1$ & CD$_{10}$ & PCD & WCD$_1$ & WCD$_{10}$ & WPCD & Data set
%\\ \hline
%0.012233 & 0.001521 & 0.090967 & 0.001037 & 0.000990 & 0.036601 & BS09  \\
%0.086695 & 0.013145 & 0.177517 & 0.000693 & 0.000676 & 0.093257 & BS16  \\
%0.198334 & 0.042751 & 0.179406 & 0.070449 & 0.008913 & 0.139242 & LSE11  \\
%0.686429 & 0.137046 & 0.226377 & 0.219336 & 0.019660 & 0.184701 & LSE15  \\
%0.436252 & 0.106502 & 0.082251 & 0.468750 & 0.016314 & 0.074255 & P08  \\  %%% Estimated
%0.693260 & 0.277417 & 0.145626 & 0.691398 & 0.048627 & 0.107775 & P10  \\
%0.003687 & 0.000199 & 0.021399 & 0.000007 & 0.000212 & 0.021416 & Int12   \\ 
%0.545241 & 0.107192 & 0.141328 & 0.003014 & 0.004837 & 0.145580 & Mult3G  \\
%0.424648 & 0.152253 & 0.159689 & 0.002346 & 0.003506 & 0.097627 & Mult3D  \\ \hline 
%\end{tabular}
%\caption{Optimal KL values for the different problems
%  described in the text.}
%\label{table_KL}
%}
%\end{center}
%\end{table*}

%% In order to deal with the target probabilities described above, a
%% valid but computationally expensive scheme would be to replicate the
%% training examples according to their target probabilities. Instead, we
%% included the probabilities in the algorithm by multiplying the
%% contribution of every example by its target probability every time the
%% example, or its reconstruction, is used in the positive or negative
%% phase, respectively.
%% {\bf ??? Not exactly (ask me why!)}

%%% BS16 - Burchuflu
\begin{figure*}[t!]
  \begin{center}
    \includegraphics[width=0.8 \textwidth]{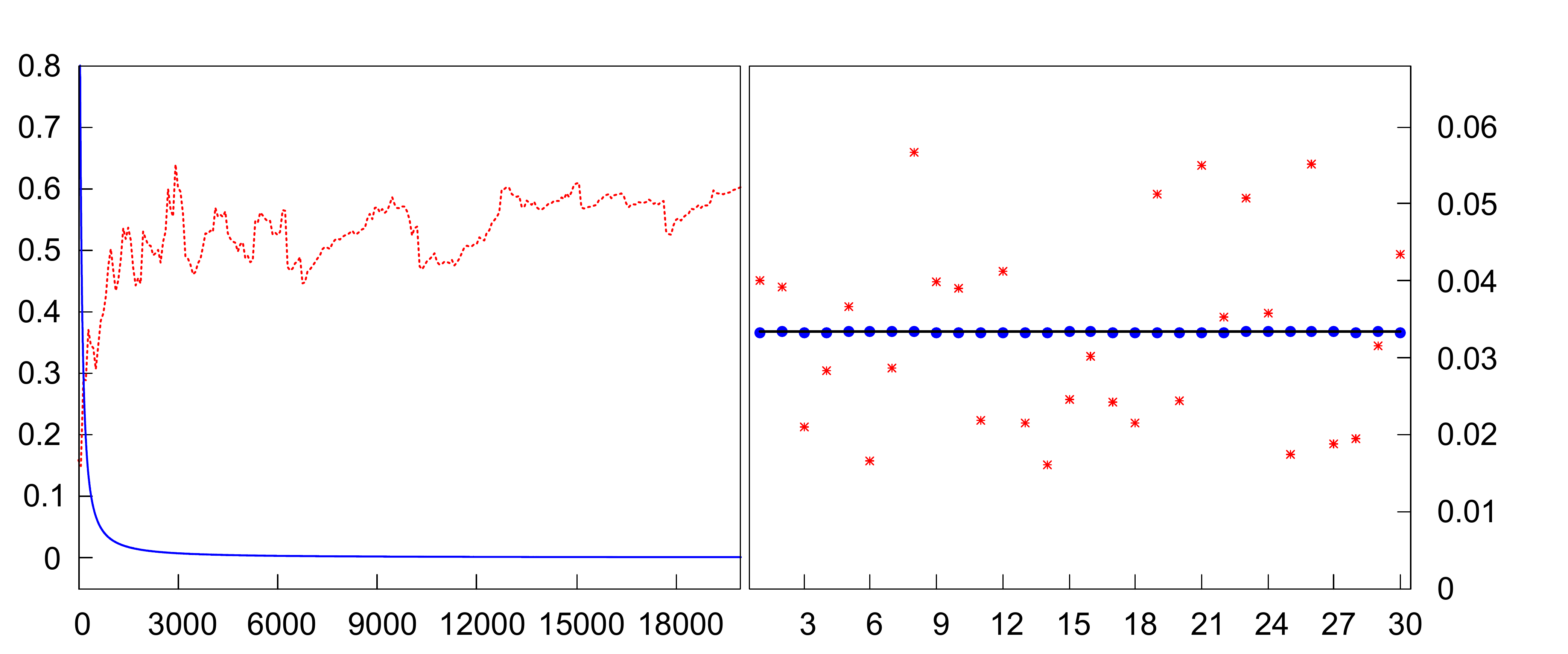} 
    \caption{KL divergence during learning (left panel) and optimal
      probabilities (right panel) of the models found by CD$_{1}$ (red
      dashed line and stars) and WCD$_{1}$ (blue solid line and
      bullets) for the BS16 data set. The $x$-axis on the left panel
      accounts for the number of epochs$/50$. The $x$-axis on the
      right panel corresponds to an integer index labelling each state
      in the training set.  The target probabilities are shown with a
      black line.}
    \label{fig_BS16_CD1WCD1}
  \end{center}
\end{figure*}

Networks were trained with standard gradient ascent for the
BS09, BS16, LSE11, LSE15, P08 and P10 problems, and stochastic
gradient ascent (with a batch size of 100) for the Int12, Mult3G and
Mult3D data sets. Weights were initialized with a Gaussian distribution
of zero mean and a variance that was suitably selected for every
model. No weight decay was used. Every network was trained for $10^6$
epochs in the BS09, BS16, LSE11, LSE15, P08 and P10 cases, and for
$10^5$ epochs in the Int12, Mult3G and Mult3D problems.

% \subsubsection{Model Selection for CD$_k$}

In order to find the optimal CD$_k$ parameters, we performed a grid
search by varying the following values:
\begin{itemize}
\item Number of hidden units: $N_v$, $2N_v$, $3N_v$, $4N_v$ and
  $5N_v$, where $N_v$ is the number of visible units.
\item Variances of the initial Gaussian weights: $1.0$, $0.1$, $0.01$,
  $0.001$ and $0.0001$.
\item Learning rates: $0.1$, $0.01$, $0.001$, $0.0001$ and $0.00001$.
\end{itemize}

Momentum was set to $0.9$.
%%% NeuronType = 1
An optimal combination of parameters was selected for every $k$ and
every problem, as explained next. Every configuration of parameters
was tested $10$ times with different random seeds. Therefore, $1250$
experiments were run for every $k$ and every data set. Out of these
experiments, we selected the combination of parameters that achieved
the smallest KL at any step of the learning process.

% \subsubsection{Model Selection for PCD}

A similar model selection was performed for PCD. The differences are
described in the following. First, since we observed that for CD$_k$
the smallest KL values were usually obtained with $5N_v$ hidden units,
only this value was tested. Second, the learning rates values
tested spanned the range from $10^{-1}$ to $10^{-8}$ in powers of 10 in
two different schemes, fixed and linearly decaying.
%% were: $0.1$, $0.01$, $0.001$, $0.0001$, $0.00001$, $0.000001$,
%% $0.0000001$ and $0.00000001$
Third, since momentum is a relevant parameter of PCD, we have tested
models with momentum values set to $0.9$ and to $0.0$.  In the end, and
as for CD$_k$, the optimal parameters were chosen as those that
achieved the smallest KL at any step of the learning process.

%%% LSE15 - Burchuflu
\begin{figure*}[t!]
  \begin{center}
    \includegraphics[width=0.8 \textwidth]{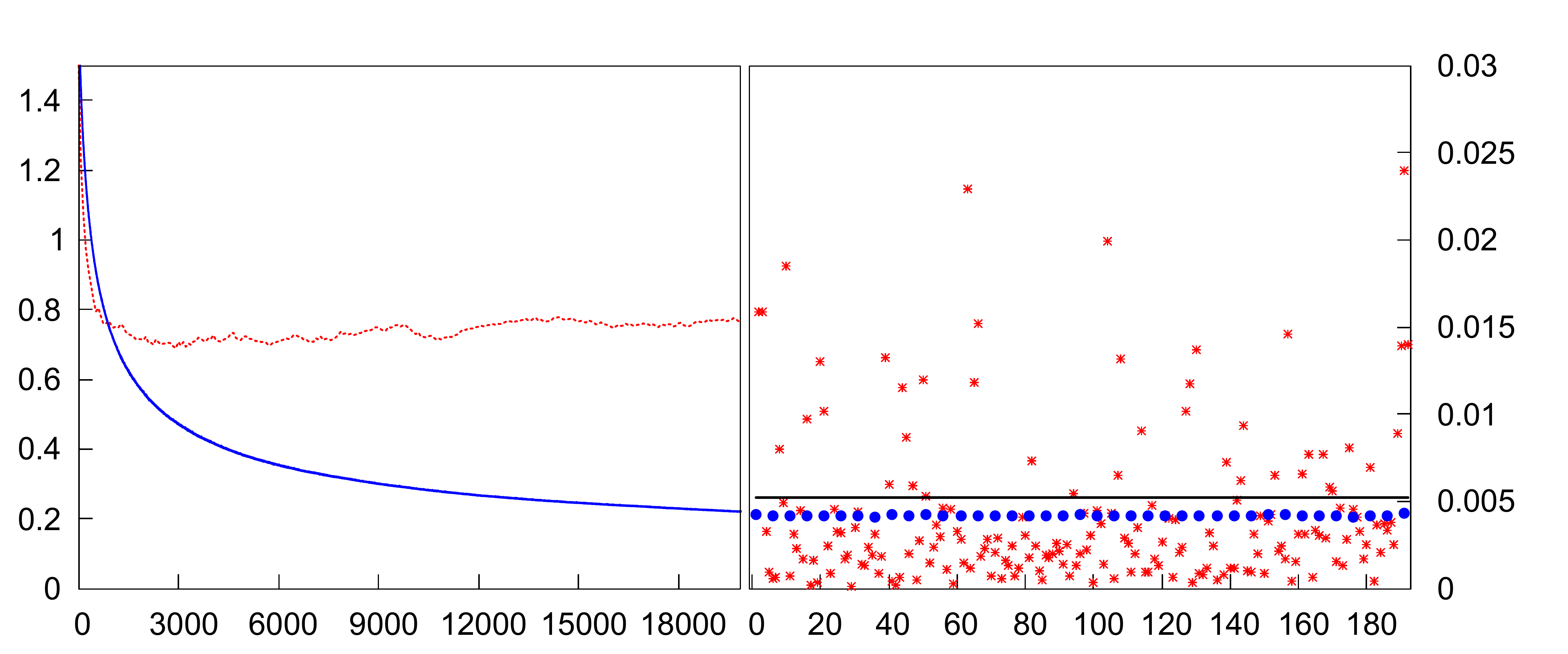} 
    \caption{Same as in figure~\ref{fig_BS16_CD1WCD1} for the LSE15 problem.}
    \label{fig_LSE15_CD1WCD1}
  \end{center}
\end{figure*}

After selecting the parameters for CD$_1$, CD$_{10}$ and PCD, the
final models were obtained for every data set in similar
conditions. To that end, we tested CD$_1$ and WCD$_1$ with the
parameters selected for CD$_1$. Furthermore, each experiment was
performed varying the number of hidden units in $\{N_v, 2N_v, 3N_v,
4N_v, 5N_v\}$, where $N_v$ is the number of visible units, and
repeated 10 times with different random seeds. The same
procedure was applied to CD$_{10}$ and WCD$_{10}$ with the parameters
of CD$_{10}$ and to PCD and WPCD with the parameters of PCD,
respectively. Notice that, while the results for the weighted models
may not be optimal, only better results can be achieved when their
parameters are specifically optimized.

%%% Parity10 - Burchuflu
\begin{figure*}[t!]
  \begin{center}
    \includegraphics[width=0.7
      \textwidth]{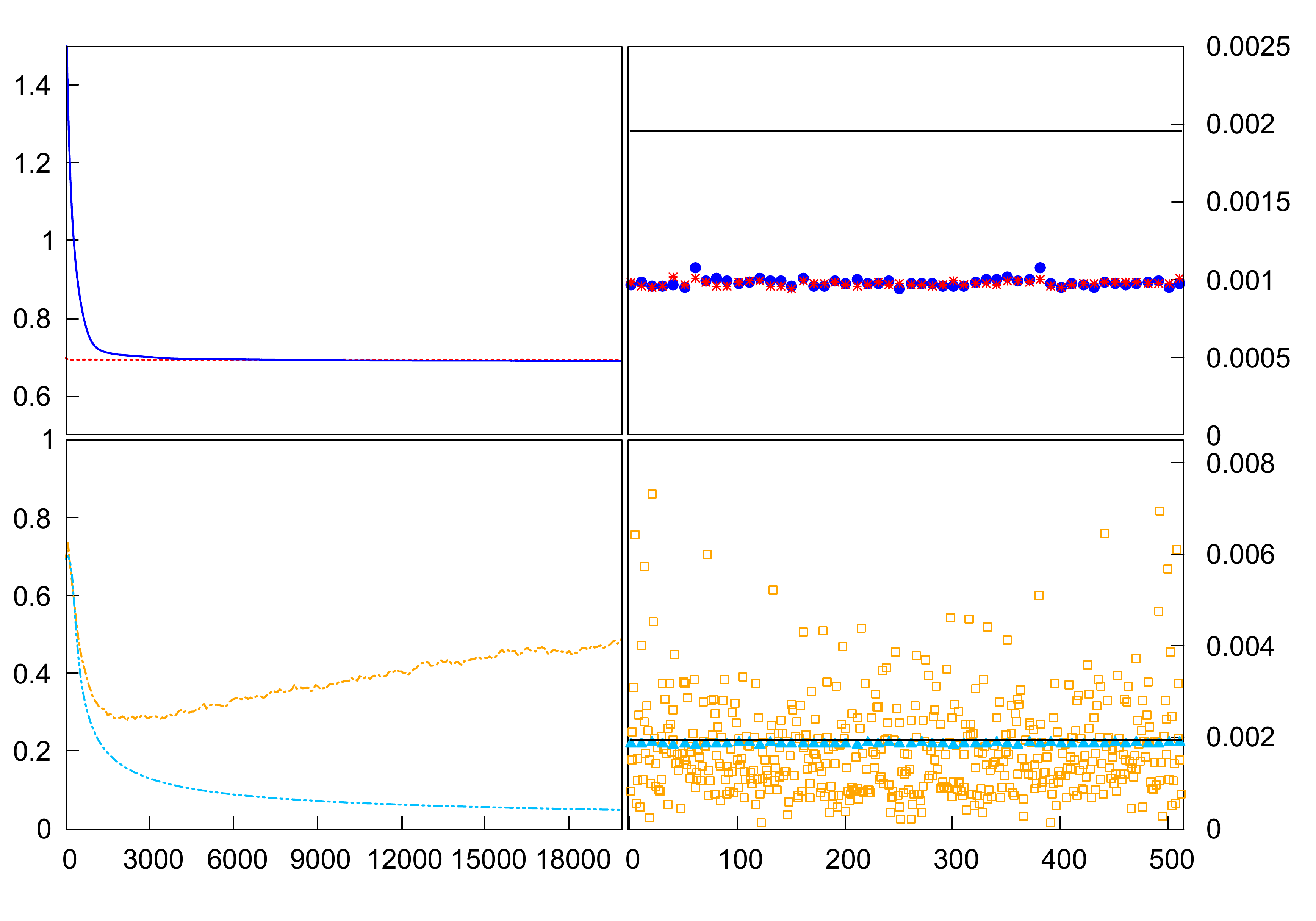}
    \caption{The top panels are the same as in
      figure~\ref{fig_BS16_CD1WCD1} for the P10 problem. The bottom
      panels are the same as in figure~\ref{fig_BS16_CD1WCD1} for the
      P10 problem in CD$_{10}$ (orange dot-dashed line and squares)
      and WCD$_{10}$ (light blue dot-dot-dashed and triangles).}
    \label{fig_Parity10_CD1WCD1CD10WCD10}
  \end{center}
\end{figure*}

% \subsubsection{Final Weighted models}

%% Finally, we tested WCD$_1$, WCD$_{10}$ and WPCD with the parameters
%% selected for CD$_1$, CD$_{10}$ and PCD, respectively.  Furthermore,
%% each experiment was performed varying the number of hidden units in
%% $\{N_v, 2N_v, 3N_v, 4N_v, 5N_v\}$, where $N_v$ is the number of
%% visible units, and repeated 5 times with different random seeds.
%% % We used the same random seeds for the different models.
%% Notice that, while these values may not be optimal, only better
%% results can be achieved when these are specifically optimized.

\subsection{Results for the Whole Training Space}

In the following we show results for the models and data sets
described above when trained with the whole training
space. Table~\ref{table_KL} summarizes
statistical averages over the 10 repetitions of each run, together
with the standard deviations. In the table we report
% the best results obtained in each case, showing the minimal
the mean KL obtained along the learning
processes. As it can be seen, in general
% when a good model is achieved,
the Weighted version of each algorithm performs better than its
non-weighted counterpart, in some cases the differences being
significantly large.  Notice that, as explained above, the learning
parameters for WCD$_k$ and WPCD have not even been optimized.
Furthermore and as will be shown in the figures, the minimum KL in the
CD and variant models is achieved at an early stage of the learning
process, but afterwards degenerates. In contrast, the evolution of the
KL in the {\em Weighted} versions is much more smooth, and the minimum
KL is attained at the end of the training. Notice also that, overall,
WCD$_{10}$ is the best performer, leading always to small KL values
that are reflected in good probabilistic models. 

%%% Multiples of 3 = 0, 3, 6 ... - Burchuflu
\begin{figure*}[t!]
  \begin{center}
    \includegraphics[width=0.8
      \textwidth]{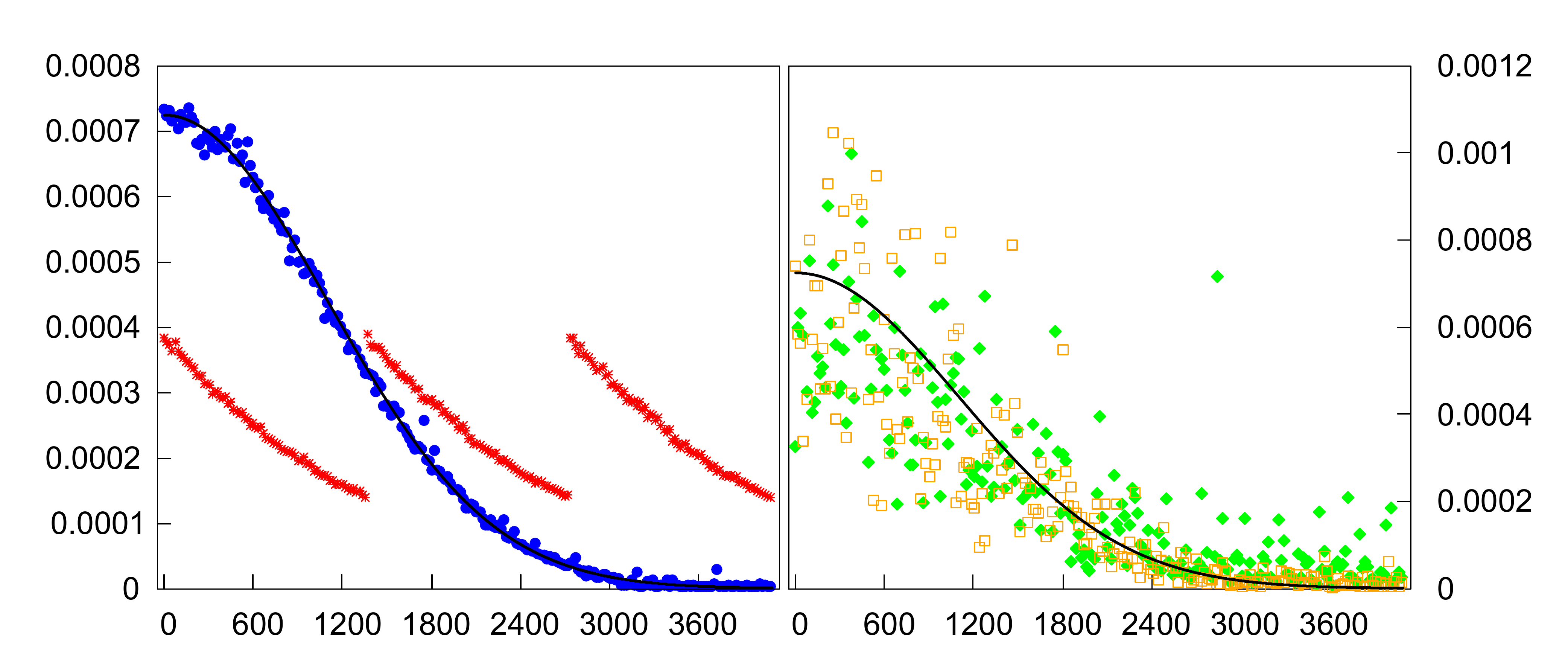}
    \caption{Optimal probabilities of the models found by CD$_{1}$
      (red stars on the left panel), WCD$_{1}$ (blue bullets on the
      left panel), CD$_{10}$ (orange squares on the right panel) and
      PCD (green diamonds on the right panel) for the Mult3G data
      set. The target probabilities are depicted as a black line. The
      $x$-axis on both panels is the index of every element of the
      data set in the corresponding ordered list (see text)}
    \label{fig_Mult3G_CD1WCD1CD10PCD}
  \end{center}
\end{figure*}

%% Notice that most of the calculations have been
%% obtained with the optimal learning parameters of the corresponding CD
%% versions (meaning that the parameters used in WCD$_1$ are the {\em
%%   optimal} ones for CD$_1$, etc). However, finding the best parameters
%% for the Weighted versions delivers even better KLs. For the sake of
%% comparison we report in the fifth and sixth rows of the table the KLs
%% ....  {\bf ???  decidir como decimos esto}

%In general the figures show the evolution of the KL
%along the learning process (on the left) and the optimal probabilities
%obtained in each case, taking as optimal the ones with minimal KL.

Since we are interested in the best probability distributions, we
report in the following figures the probabilities of the data in the
training space for the best models
out of the 10 repetitions performed in each case.
% obtained
These probabilities are compared with the
corresponding target ones (on the right). We also report the KL values
during the training process (on the left) as a way to evaluate the
evolution of the models found during learning.
Figure~\ref{fig_BS16_CD1WCD1} shows the CD$_1$ and WCD$_1$ results
obtained for the BS16 problem. As it can be seen, while the CD$_1$
probabilities are not dramatically wrong, the comparison between
WCD$_1$ and the target probabilities is outstanding. Regarding the KL,
not only the optimal and asymptotic values are much better in WCD$_1$,
but also the models found are much more stable, according to the small
local variability of the WCD$_1$ KL curve.

%%% yet another one :(
\begin{figure*}[t!]
  \begin{center}
    \includegraphics[width=0.8 \textwidth]{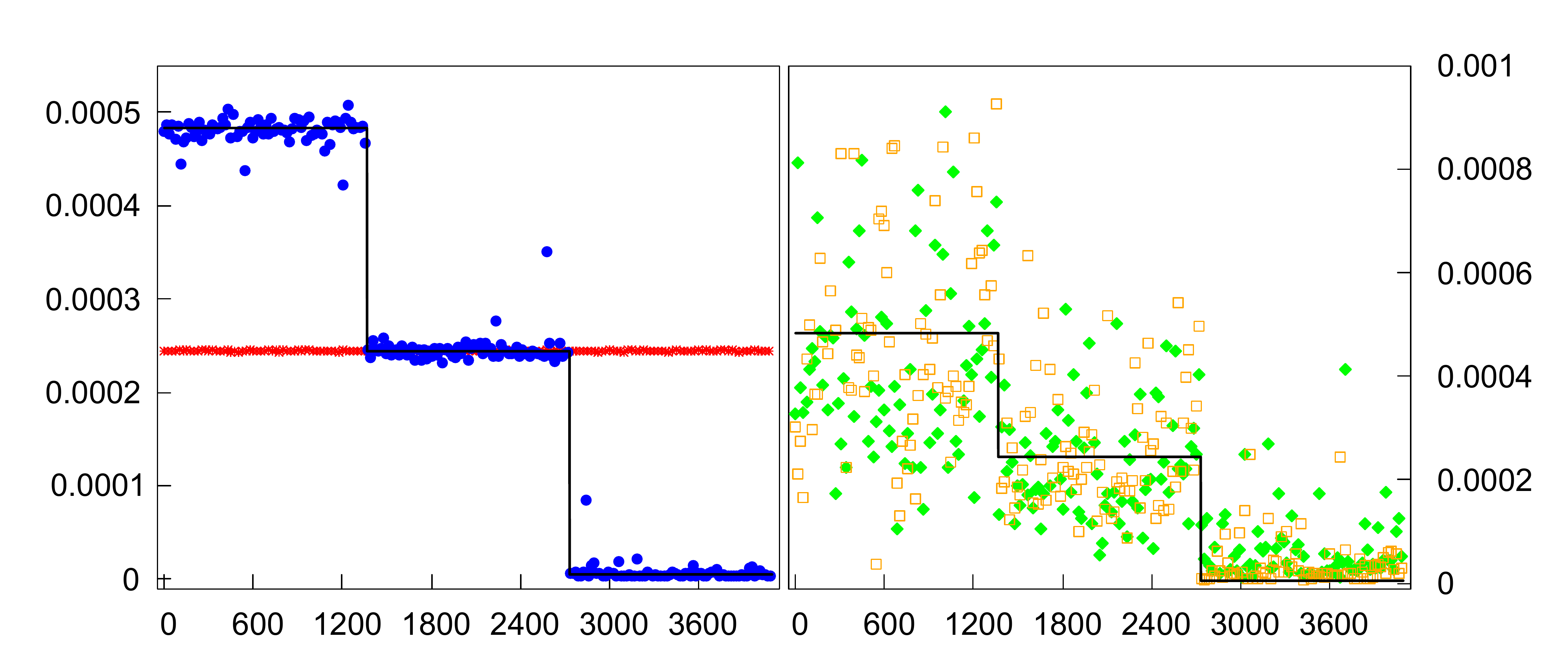} 
    \caption{Same as figure~\ref{fig_Mult3G_CD1WCD1CD10PCD} but for the data set Mult3D}
    \label{fig_Mult3D_CD1WCD1CD10PCD}
  \end{center}
\end{figure*}

Figure~\ref{fig_LSE15_CD1WCD1} shows the same quantities for the LSE15
problem. In this case the evolution of the KL in CD$_1$ is much more
smooth than in the previous case, though its optimal value found is
much worse than the one found in the BS16 problem. Once again, the
optimal WCD$_1$ KL is much lower than the corresponding CD$_1$ one,
pointing to a better probabilistic model when compared to the target
distribution.  In the same way, the WCD$_1$ KL performs roughly as in
the BS16 case, though convergence to the asymptotic value is much
slower, while it still keeps its decreasing behavior. All these
features are reflected in the optimal probabilities reported on the
right panel.  As it can be seen, the WCD$_1$ probabilities are much
closer to the target ones than those generated by CD$_1$, being also
much more uniform. Furthermore, the CD$_1$ probabilities present much
larger oscillations, and in particular there are several outlayers
that take a large amount of the probability mass individually, as
mentioned in the introduction. In this case, however, though the
WCD$_1$ algorithm produces almost uniform probabilities as desired,
the exact value is not reproduced as in the BS16 case, meaning that
states not contained in the training space acquire non-zero
probabilities. This does not happen in the WCD$_{10}$ estimate, which
achieves a very small KL value and therefore fits very well the target
probabilities. As a matter of fact, this also happens in the BS16
problem, where the model probabilities already sum up to $1$.  The
smaller version of the same problems (BS09 and LSE11) show similar
behavior.

%%% Mult3Gauss_TestProbs - Burchuflu
\begin{figure*}[t!]
  \begin{center}
    \includegraphics[width=0.8 \textwidth]{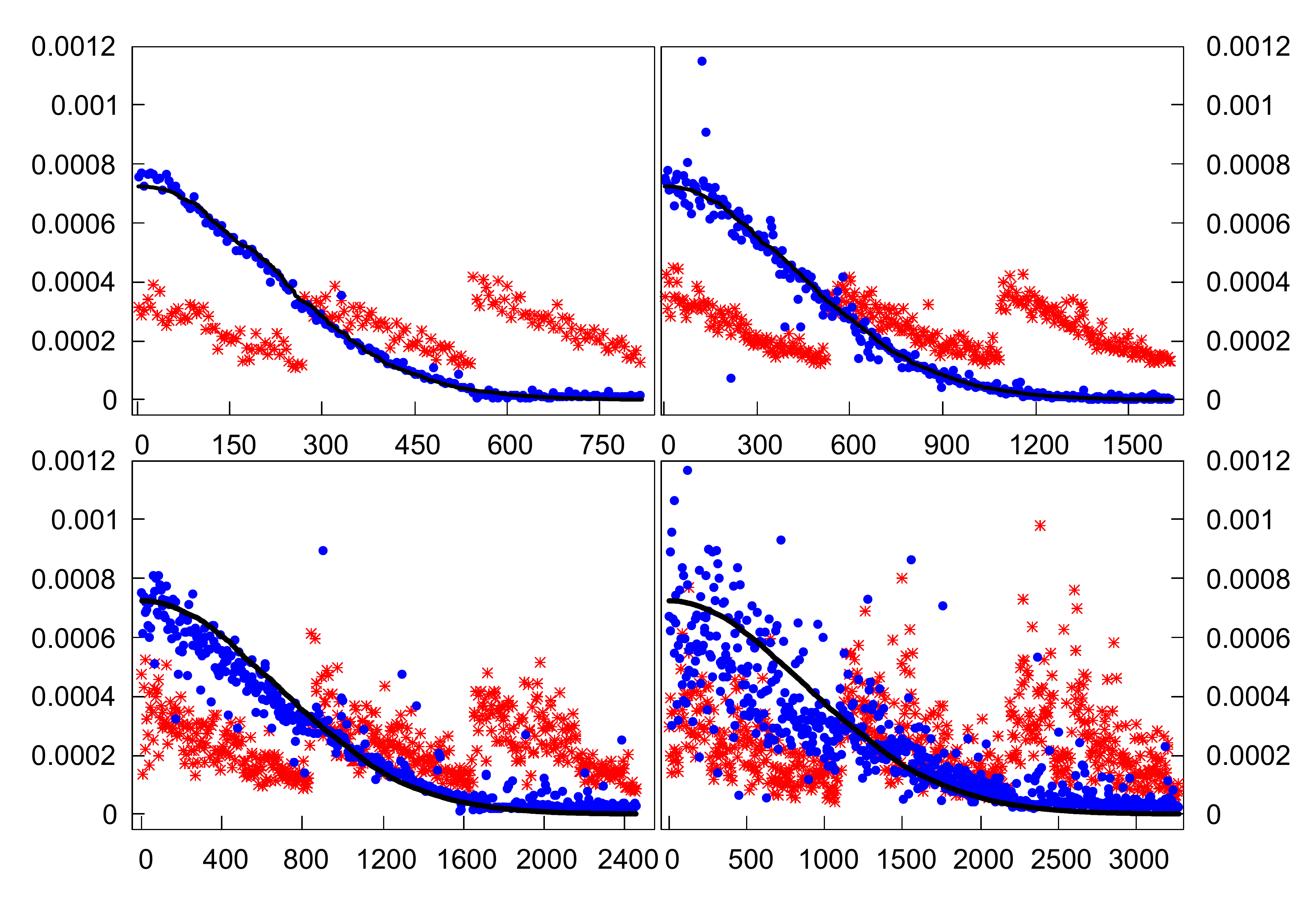} 
    \caption{Probabilities of the test set corresponding to the models
      with minimum KL in the training set found by CD$_{1}$ (red stars) and
      WCD$_{1}$ (blue bullets) for the Mult3G problem. Top Left: 80\%/20\%
      training/test. Top Right: 60\%/40\% training/test. Bottom Left:
      40\%/60\% training/test. Bottom Right: 20\%/80\%
      training/test. The $x$-axis are similar to
      figure~\ref{fig_Mult3G_CD1WCD1CD10PCD}, but only the examples in
      the test set are shown. The target probabilities are shown with a
      black line.}
    \label{fig_Mult3Gauss_TestProbs_CD1WCD1}
  \end{center}
\end{figure*}

Figure~\ref{fig_Parity10_CD1WCD1CD10WCD10} shows results for the
toughest problem analyzed in this work, the P10 data set. The upper
and lower plots show the (W)CD$_1$ and (W)CD$_{10}$ results,
respectively. In this case neither CD$_1$ nor WCD$_1$ are able to
learn a sensible model, maybe because of the difficulty of the
problem. Remarkably, the optimal CD$_1$ KL is almost identical to the
optimal WCD$_1$ KL, and the evolution of the KL along learning in both
cases is very similar and smooth but poor. The consequence of all this
is that the resulting probability distributions are almost identical,
none of them making much sense. Notice that while the probabilities
found are quite uniform, they are still far away from the real target
values: the training space, consisting in half the total space,
receives half the total probability while it should sum up to 1. A
different situation is found for the CD$_{10}$ and WCD$_{10}$
predictions, as shown in the lower plots in the same figure. In this
case the CD$_{10}$ KL finds a minimum but degenerates afterwards,
while in the WCD$_{10}$ case it behaves as in the previous problems,
monotonically decreasing and approaching its asymptotic value, which
is much lower than the CD$_{10}$ one. The resulting probabilities are
closer to the target ones in both cases, but it is remarkable how
better WCD$_{10}$ performs. While the resulting CD$_1$ and WCD$_1$
models are very similar, going from $k=1$ to $k=10$ leads to a much
more accurate model in the WCD$_{10}$ case, whereas CD$_{10}$ produces
once again a highly non-uniform distribution with scattered values in
a broad range.
%So you are amazed at how chachiful WCD10 is, isnt it?

In Fig.~\ref{fig_Mult3G_CD1WCD1CD10PCD} we report in two panels our
results for the Mult3G problem. We have split them in two because of
the different scales of the resulting probability distributions.  The
left panel shows CD$_1$ and WCD$_1$, while the right panel depicts the
probabilities obtained in CD$_{10}$ and PCD.  As expected, the quality
of the Weighted version is markedly better than the standard CD$_1$ 
prediction, the later being clearly unable to reproduce the target
distribution. In the same token, WCD$_1$ performs better than
CD$_{10}$ and PCD. It is important to take into account that the
ordering of the states in the Mult3G problem is relevant. Three
different classes have been built, the first one containing the states
corresponding to the values $0,3,6,\ldots$ in this order, the second
one containing the values $1,4,7,\ldots$ in this order, and so on. As
it can be seen, CD$_1$ detects that lower values have larger
probabilities, but it is not able to discern among the different
classes.  In the CD$_{10}$ and PCD cases, both estimations of the
distribution probabilities capture the main trends of the target
probabilities. However, in both cases fluctuations around the right
values are large and comparable, maybe PCD performing a little bit
worse, in agreement with the KL values reported in the table.  On the
other hand, WCD$_1$ once again recovers a nice model, clearly
discriminating among the three groups.

Finally, in Fig~\ref{fig_Mult3D_CD1WCD1CD10PCD} we report the
probability distributions for the Mult3D problem obtained in CD$_1$,
WCD$_1$, CD$_{10}$ and PCD as in
Fig.~\ref{fig_Mult3G_CD1WCD1CD10PCD}.  In this case the WCD$_1$
prediction is dramatically better than the CD$_1$ one, as the later is
only able to assign essentially the same probability to each state,
not being able to discern any feature of the problem. In contrast,
WCD$_1$ performs well and clearly discriminates the three categories
of the problem, with quite uniform probability in each group. In
We also see that CD$_{10}$ and PCD performs similarly to the
Mult3G case.

%%% Mult3Uniform_TestProbs - Burchuflu
\begin{figure*}[t!]
  \begin{center}
    \includegraphics[width=0.8 \textwidth]{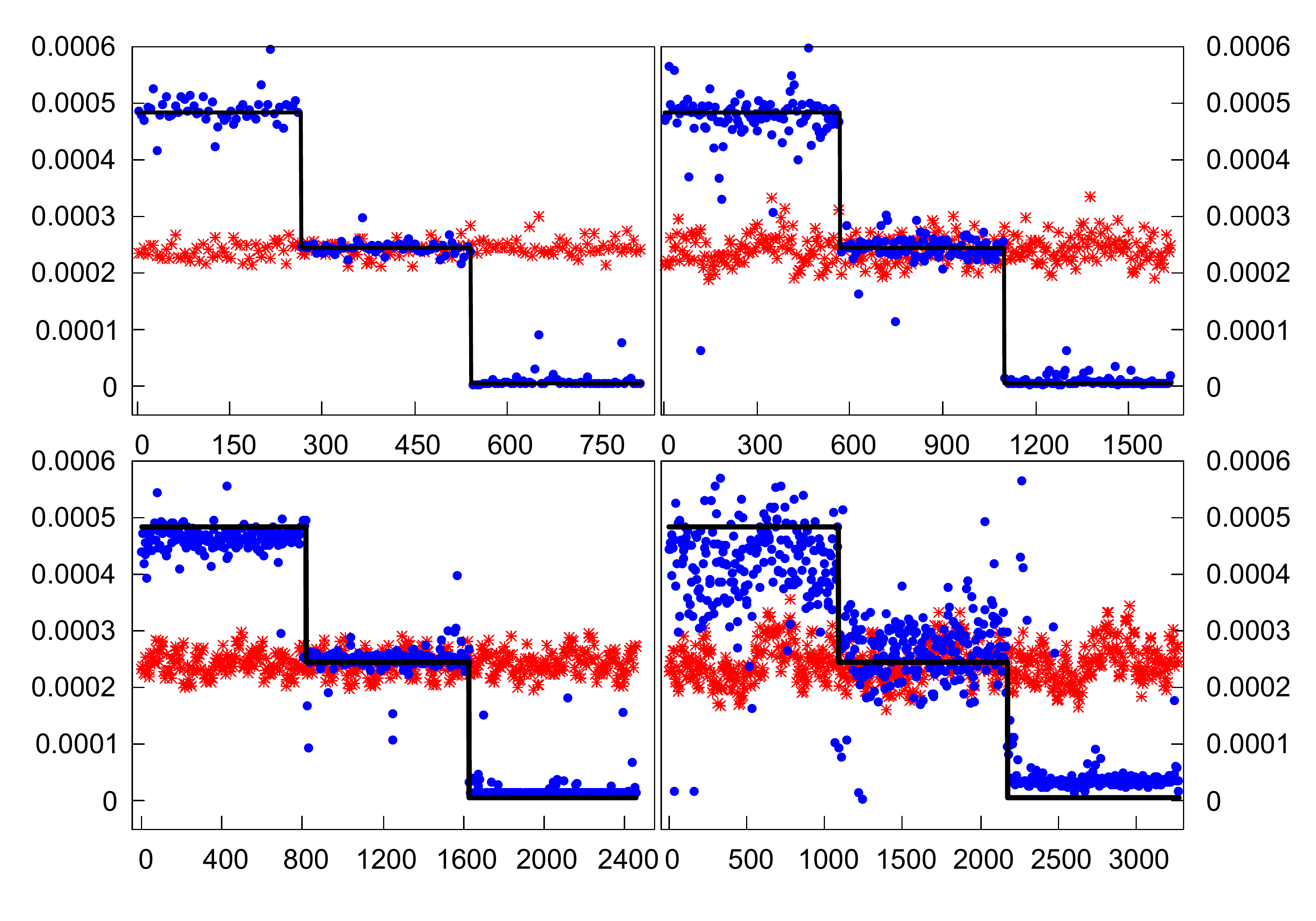} 
    \caption{Same as in figure \ref{fig_Mult3Gauss_TestProbs_CD1WCD1} for
      the Mult3D data set.}
    \label{fig_Mult3Uniform_TestProbs_CD1WCD1}
  \end{center}
\end{figure*}

\subsection{Generalization Results}

The experiments described in the previous section were focused on
trying to obtain the models with minimum KL in order to fit the
probabilities of the training space. Therefore, they were analyzing
the approximation capability of the respective models. In this section
we focus on the generalization capability, which in many cases is the
most important objective of the system.

To that end, we performed a number of experiments with the Int12,
Mult3G and Mult3D data sets. The training set in these cases was
taken to be a fraction of the whole training space used in the
previous section, leaving the rest of states for the test set.  The
fraction values used for the training/test sets were 80\%/20\%,
60\%/40\%, 40\%/60\% and 20\%/80\%. The model with the minimum KL in
the training set was saved and subsequently tested on the test set. As
in the previous experiments, we compared standard CD$_1$, CD$_{10}$,
PCD, and their {\em Weighted} counterparts, WCD$_1$, WCD$_{10}$ and
WPCD. The parameters of the models were those already selected in the
experiments performed with the complete training space.

Figure~\ref{fig_Mult3Gauss_TestProbs_CD1WCD1} depicts results for the
Mult3G problem. In essence, both CD$_1$ and WCD$_1$ keep the same
structure observed when all states in the training space are used for
learning. That means that CD$_1$ is always missing the main trends of
the target probability, while WCD$_1$ keeps up fairly well. Still, the
quality of WCD$_1$ worsens when the fraction of states used for
training is reduced, as expected. But it is remarkable that, in all
these cases, WCD$_1$ is able to generalize successfully. In particular,
we notice that WCD$_1$ trained with just a 20\% of the complete
training space performs much better that bare CD$_1$ trained with the
whole training space. Finally,
figure~\ref{fig_Mult3Uniform_TestProbs_CD1WCD1} presents the same
quantities as in the previous figure, for the Mult3D problem. As
discussed above, this is a hard problem for CD$_1$ as it is never able
to capture any single feature of the data. In contrast, WCD$_1$ holds
up quite well even under severe training space restrictions.  Similar
results are found for the rest of the models and data sets when the
training space is reduced as above.

%% ??? Decidir: Mostrar la KL *SUMADA* de Train y Test o solo la parte de
%%     cada uno y explicar los valores negativos

\section{Experiments with large size data sets}
\label{section-experiments-Parzen}

\begin{figure*}[t!]
  \begin{center}
    \includegraphics[width=0.8 \textwidth]{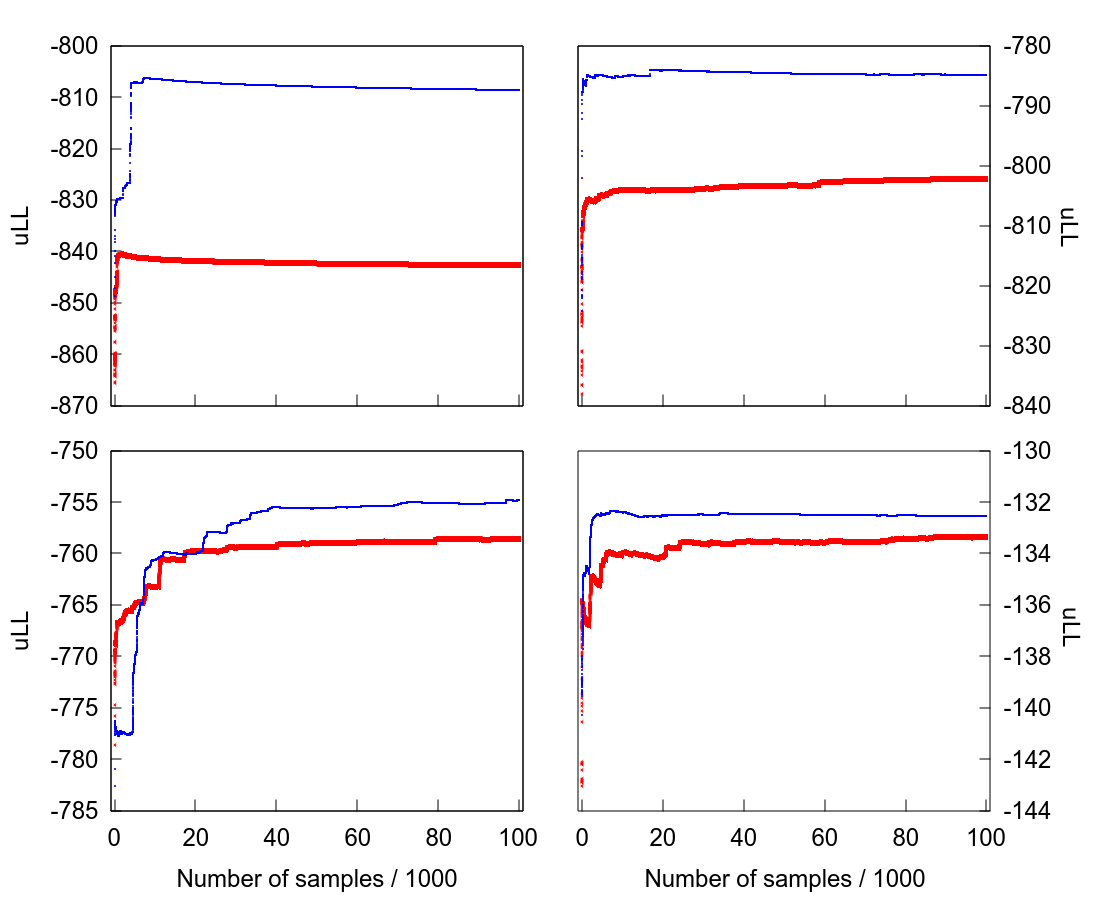} 
    \caption{Comparison of the quality of the samples generated using
      a Parzen window procedure for WPCD (blue curves) and PCD (red
      curves) in for the Caltech101 (upper left), Fashion-MNIST
      (upper right), MNIST (lower left) and OCR-Letters (lower right)
      problems. The x-axis represents the number of samples generated
      by each model and used to compute the uLL.}
    \label{fig_Parzen_windows}
  \end{center}
\end{figure*}

In this section we show how the WCD technique performs in real-world,
high dimensional data sets where an exact computation of the
probability of each state is unfeasible. Since the evaluation of the
likelihood is not possible, we employ an alternative estimation of the
quality of the results by using a Parzen window
estimator~\cite{Breuleux_Parzen_2011} of the probabilities of the test
set, obtained from a set of samples performed over the learned model.
For computational reasons, we do not use the AIS algorithm to estimate
the probabilities, which, in addition, is very hard to validate in
really high dimensional spaces as the ones considered here.  In short,
our Parzen window estimator uses a Gaussian probability density
distribution $g({\bf x};{\bf x}_i,\sigma_i)$ centered at each point
${\bf x}_i$ of the sample set and with standard deviation $\sigma_i$
(which we take to be the same for all points), as
in~\cite{Breuleux_Parzen_2011}. The idea then is to assign a
probability density at each point ${\bf y}_j$ in the test set equal to
the averaged sum of $g({\bf y}_j;{\bf x}_i,\sigma_i)$ over all
samples. One then uses these averaged probability densities to build
an unnormalized log-likelihood (uLL) of the test set, which depends on
the set of samples used. More specifically, we assign to each point of
the test set ${\bf y}_j$ an averaged Gaussian value
\[
G({\bf y}_j) = {1\over N_s} \sum_i g({\bf y}_j; {\bf x}_i,\sigma_i) 
\]
where $N_s$ is the number of samples employed, and
\[
{\rm uLL} = {1\over N_t}\sum_j \ln G({\bf y}_j) \ ,
\]
with $N_t$ the number of members of the test set.
For the sake of comparison, we perform this procedure twice, using a
maximum of $10^5$ samples generated by the model obtained with WPCD
and PCD, respectively.

\begin{table*}[t!]
\centering
\caption{Details of the data sets employed in the experiments}
\label{Table:DatasetDescription}
\begin{tabular}{|l|l|l|l|}
\hline
Dataset            & Train size & Test size & Num. of features \\ \hline
Caltech101 Silhouette & 4100   & 2307      & 784              \\ \hline
OCR-Letters        & 32,152     & 10,000    & 128              \\ \hline
MNIST              & 60,000     & 10,000    & 784              \\ \hline
Fashion-MNIST      & 60,000     & 10,000    & 784              \\ \hline
\end{tabular}
\end{table*}

We have tested the proposed approach on the following four data sets:
MNIST, Fashion-MNIST, Caltech101 silhouettes and OCR letters.
The MNIST data set is a well known benchmark problem corresponding to
$28 \times 28$ grayscale images of hand-written digits.  The fashion
MNIST contains $28\times 28$ grayscale images, associated with 10
different clothing categories (dress, coat, shirt, $\ldots$). The
CalTech101 Silhouettes data set contains $28\times 28$ binary images,
containing items from 101 different categories (faces, leopards, ants,
butterflies, $\dots$). Finally, the OCR-Letters data set contains
$16\times 8$ samples of grayscale images of handwritten letters.
Table \ref{Table:DatasetDescription} shows the number of features of
each data set and number of samples in the train and test partitions.

In both the WPCD and PCD cases, the architecture of the network contained
a variable number of visible units (depending on the problem) and a
fixed number of hidden units, which was set to 500.  In all cases, the
networks were trained for 1000 epochs with stochastic gradient ascent
and a batch size of 100 examples. Different values of the learning
rate have been tested, in a logarithmic mesh spanning the range
$[0.001, 1.0]$ to find the optimal value in each case. The momentum
and the weight decay factors were set to 0.75 and 0.0002,
respectively. Finally, the learning rate followed a linearly decaying
scheme. Other combination of parameters have also been tested, to find
that the best values lay in the mentioned ranges. In all cases, the
selection criterion was set to get the highest test accuracy in
supervised models were the labels were appended to the training
examples with a one-hot coding representation.
The final models were trained in an unsupervised way with the optimal
values found with the model selection described above.

Figure~\ref{fig_Parzen_windows} shows the comparison of the uLL of the
test set as a function of the number of samples employed in the Parzen
window estimator, for the four data sets analyzed. In all cases, the
blue and red curves correspond to the results obtained in WPCD and
PCD, respectively.  The upper left and right panels show the uLL for
the Caltech101 and Fashion-MNIST, while the lower left and right
panels correspond to the MNIST and OCR-Letters problems. As can be
seen, in all cases WPCD produces higher uLL values, pointing to a
better model (in a Parzen window sense) of the learned model with
respect to the test set. In must be kept in mind, however, that the
uLL is not a real estimation of the log-likelihood of the data, and
that in all cases there is an arbitrary constant that sets the origin
on the scales.  That means that only the relative differences between
two estimations make sense. Anyway, higher scores can be attributed to
better models, although it is not possible to quantify how better.
Notice that there is always a transient regime at the beginning of the
curves where the variation of the uLL is large, corresponding to a poor
statistical representation produced by the reduced number of samples
employed. However, as this number increases, the curves approach a
more stable regime where the predictions seem to stabilize, with WPCD
approaching a better statistical representativity with a smaller
number of samples.

\section{Conclusions}

In summary, in this work we propose a variant of the standard CD
learning algorithm for RBMs that modifies the negative phase of the
gradients involved in the weights update rule. The new negative phase
is computed as a weighted average over the members of the batch, where
the weighting coefficients are the relative model probabilities in the
batch. This is a cheap modification that nevertheless delivers better
performance, both in terms of KL and optimal model probabilities. We
have tested the proposed algorithm against a set of small problems
where exact probabilities can be evaluated, to find that the
statistical representation of the learned models is much better than
the one obtained in CD and PCD. In large problems, were a direct
measure of the probabilities is unfeasible, a Parzen window evaluation
of the quality of the resulting models still indicates that WPCD
performs better that their alternatives.  In any case, it is important
to realize that, when the data can be processed and reduced to a
small-dimensional space (for example, in a feature extraction stage),
weighting the negative phase is a good choice that improves the
probability description of the model. Furthermore, the weighting
scheme can be extended to other useful techniques alternative to CD.
% such as Parallel
% Tempering~\cite{desjardins-et-al-Parallel-Tempering-2010-AISTATS}.

Possible future work involves a more sophisticated selection of
elements entering in the weighted negative phase.  This can, for
instance, involve not only members from the training set, but also
neighboring ones that, for continuity reasons, could also contain
relevant information. For small problems, comparison to exact gradient
calculations can also be carried out in order to contrast the
statistical averages of the exact calculation to the approximated one
in the learning scheme. 
Another interesting aspect is to analyze the convergence properties of
the proposed methodology. Along this line, one can try to extend the
analysis performed in~\cite{Rollo_1, Cara_Kulo_1, Rollo_3}, although the task is
not evident for the first two since they apply to continuous units and
not binary ones. 

\section*{Acknowledgments}
ER: This work was partially supported by MINECO project
TIN2016-79576-R.
FM: This work has been supported by MINECO grant
No. FIS2017-84114-C2-1-P
%FIS2014- 56257-C2-1-P
from DGI (Spain).  JD: This work was supported by the grant
TIN2017-89244-R from MINECO (Ministerio de Economia, Industria y
Competitividad) and the recognition 2017SGR-856 (MACDA) from AGAUR
(Generalitat de Catalunya).  Part of the hardware used for this
research was donated by the NVIDIA Corporation.

%%% BIBLIOGRAPHY
%%%\small
\bibliography{bibliography-paper}

\end{document}